\documentclass[12pt,draftclsnofoot,onecolumn]{IEEEtran}

\usepackage{amsmath,amssymb,amsfonts,amsthm,bm}
\usepackage{graphicx}
\usepackage{cite}
\usepackage{url}
\usepackage{algorithm,algorithmic}
\usepackage{multirow}
\usepackage{booktabs}
\usepackage{array}
\usepackage{mathtools}
\usepackage{enumitem}
\usepackage{color}
\usepackage{hyperref}
\usepackage[T1]{fontenc}
\usepackage{pifont}
\newcommand{\xmark}{\ding{55}}
\newcommand{\cmark}{\ding{51}}
\usepackage[caption=false,font=footnotesize]{subfig}

\usepackage{dblfloatfix}  
\usepackage{etoolbox}     
\usepackage{mathalfa}     
\usepackage{eqparbox}     

\def\BibTeX{{\rm B\kern-.05em{\sc i\kern-.025em b}\kern-.08em%
    T\kern-.1667em\lower.7ex\hbox{E}\kern-.125emX}}

\title{Communication-Efficient and Interoperable Distributed Learning}

\author{Mounssif Krouka and Mehdi Bennis, Fellow, IEEE
\thanks{M. Krouka and M. Bennis are with the Centre for Wireless Communications (CWC), University of Oulu, Finland (emails: \{mounssif.krouka, mehdi.bennis\}@oulu.fi). Preprint version. Submitted for peer review.}
}
\begin{document}
\maketitle
\begin{abstract}
Collaborative learning across heterogeneous model architectures presents significant challenges in ensuring interoperability and preserving privacy. We propose a communication-efficient distributed learning framework that supports model heterogeneity and enables modular composition during inference. To facilitate interoperability, all clients adopt a common fusion-layer output dimension, which permits each model to be partitioned into a personalized base block and a generalized modular block. Clients share their fusion-layer outputs, keeping model parameters and architectures private. Experimental results demonstrate that the framework achieves superior communication efficiency compared to federated learning (FL) and federated split learning (FSL) baselines, while ensuring stable training performance across heterogeneous architectures.
\end{abstract}
\vspace{-0.2cm}
\section{Introduction}
\vspace{-0.1cm}
Driven by advances in wireless technologies, the number of interconnected devices has grown rapidly, leading to the generation and transmission of massive volumes of data. This surge has enabled a wide range of data-driven artificial intelligence (AI) applications \cite{8869705, 9349624, 8766143}. In centralized AI systems, where raw data is transmitted to remote servers for training, communication overhead is high, and data privacy is compromised.
To address these challenges, Federated Learning (FL) was introduced, allowing clients to keep their local data and share only model updates with a central server \cite{FL, YANG202233, MAL-083}. However, transmitting the full model updates in each training round leads to high communication costs. Additionally, FL requires all clients to adopt a common model architecture, limiting flexibility and promoting vendor lock-in, which impedes support for heterogeneous models.
To reduce communication overhead, Split Learning (SL) serves as a more efficient alternative. In SL, the client processes a portion of the model and transmits the intermediate output from a designated cut-layer to the server, which completes the remaining computation \cite{vepakomma2018splitlearninghealthdistributed, GUPTA20181, 10118601, 9685045}. As a
result, model parameters are not exchanged.

When extended to multiple clients, Federated Split Learning (FSL) enables collaborative training with distributed client-side model partitions and a shared server-side model \cite{thapa2022splitfedfederatedlearningmeets,fan2025p3slpersonalizedprivacypreservingsplit, papageorgiou2025collaborativesplitfederatedlearning}. However, FSL requires clients to rely on the server’s model part during both training and inference. This exposes server-side model parameters and architecture to the server, constraining modular AI designs and limiting component-level competition \cite{10024837}.

In this work, we propose a communication-efficient distributed learning architecture that supports heterogeneous AI models across clients, while enabling cross-vendor interoperability and model composition during inference. The clients agree on a common output dimension of a fusion-layer, which partitions each model into a lower-level base block and an upper-level modular block. This architecture enables end-to-end training without exposing model parameters, gradients, or architecture details to other clients or the server. Furthermore, modular blocks can be flexibly interchanged across clients via the fusion layer, facilitating model composition during inference. We highlight the main contributions as follows:
\begin{itemize}
    \item We propose a novel two-stage training algorithm that enables personalization of the base block using local data and generalization of the modular block using concatenated fusion-layer outputs from the clients.
    
    \item We allow multiple local updates to the base block before transmitting fusion-layer outputs, unlike FSL which performs only a single update per communication round, thereby reducing communication frequency and overhead.
    
    \item We enable heterogeneous and proprietary model designs by ensuring that the model parameters, gradients, and architecture of each client remain private and restricted to the client side, with only the output of the fusion-layer shared with the server.
    
    \item We leverage the same fusion-layer output dimension among the clients to standardize the interface between model partitions, enabling modular block interoperability across heterogeneous client models and supporting flexible multi-vendor deployments.
    
    \item We empirically demonstrate that our framework significantly reduces communication overhead compared to FL and FSL, while maintaining high accuracy. We further validate its robustness to model heterogeneity and its support for seamless modular block interoperability during inference across diverse client models.
\end{itemize}

\vspace{-0.3cm}
\section{system model}
\vspace{-0.2cm}
\label{sys_model}
  \begin{figure*}[t]
	\centering
	\includegraphics[width=1\linewidth, height = 7.9cm]{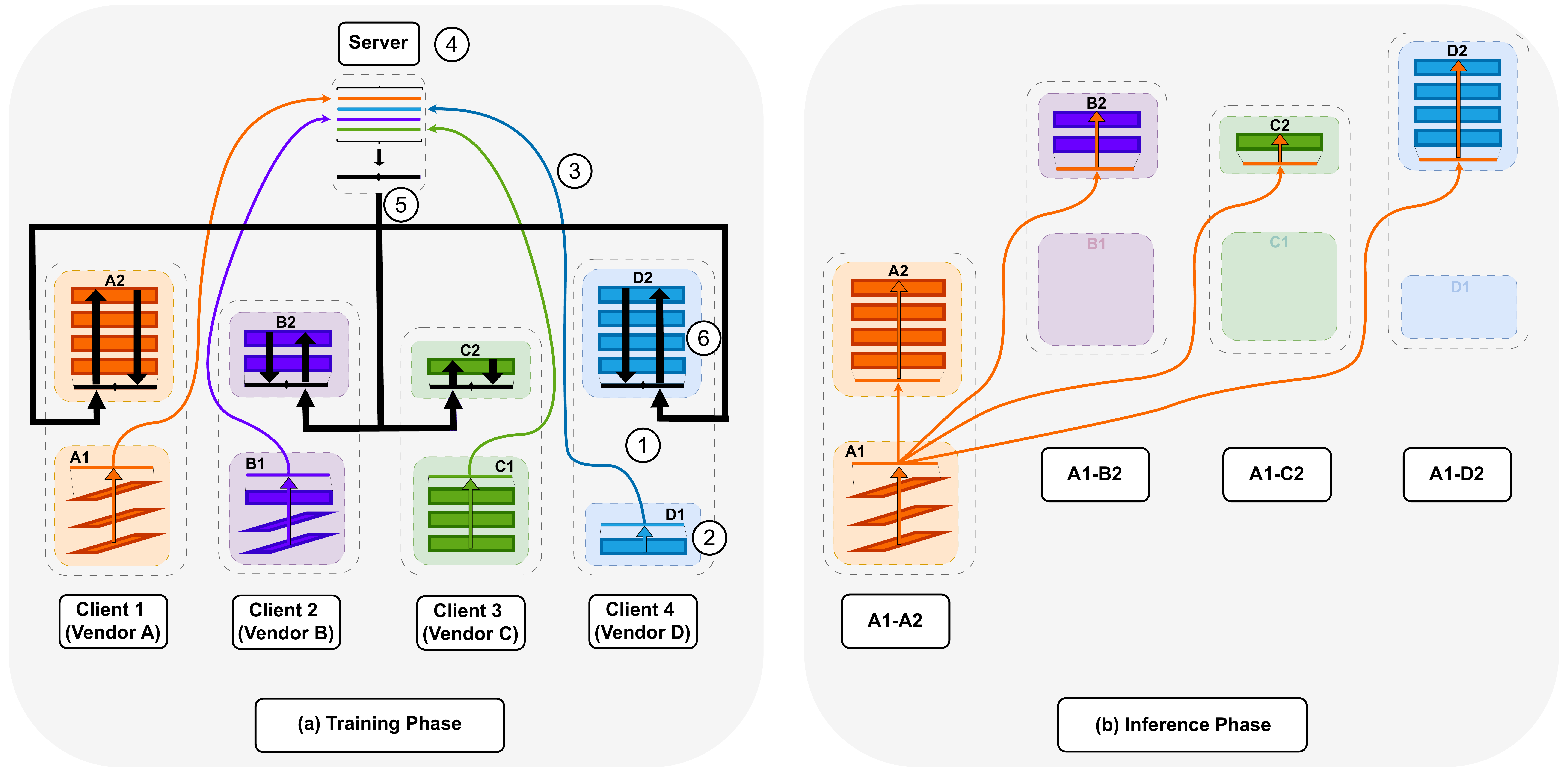}
	\caption{Overview of the proposed framework’s training and inference phases. (a) Training: local base block updates and server-side concatenation of fusion-layer outputs for modular block training. (b) Inference: base block combined with modular blocks from different clients to enable cross-client composition.} 
	\label{Fig:diagram} 
\end{figure*}
We consider a distributed learning system coordinated by a remote server with $N$ clients, each having a local dataset and a model with a client-specific architecture. 
To enable collaboration across clients while preserving model heterogeneity and ownership,
we adopt an architecture that partitions each client's model into two components: a base block and a modular block, separated by a fusion-layer with a common output dimension across all the clients.
The base block layers are from the input until the fusion-layer, and the modular block layers constitute the remaining part until the output layer.\\
\subsection{Training Phase}
The end-to-end training process is shown in Fig. \ref{Fig:diagram}a, where we assume that each client deploys a model architecture from a different vendor. We denote the base block and the modular block of a given vendor (e.g., vendor $X$) as $X1$ and $X2$, respectively. We follow the numbering in Fig. \ref{Fig:diagram}a and describe the training steps in this way:
\begin{itemize}

\item Step 1: Each client performs $\tau$ update steps to the base block parameters.

\item Step 2: Clients start forward propagation on the base block until the fusion-layer.

\item Step 3: Each client sends its fusion-layer output and labels to the server.

\item Step 4: On the server side, the server concatenates the outputs received from the clients.

\item Step 5: The server broadcasts the result to the clients.

\item Step 6: Each client uses the data received from the server as input to the modular block and updates its parameters.

\end{itemize}
\subsection{Inference Phase} 
After training is complete, each client retains a personalized base block trained on local data and a generalized modular block trained using the concatenated outputs of the fusion layers of all clients. A common fusion-layer output dimension is required across clients to ensure compatibility and support modular composition.
This standardization enables vendors to design model architectures independently, without concern for cross-vendor compatibility. Fig. \ref{Fig:diagram}b illustrates an example of modular composition, where the base block of vendor $A$, referred to as $A1$, is deployed with the modular blocks $A2$, $B2$, $C2$, and $D2$, which correspond to the vendors $A$, $B$, $C$, and $D$, respectively. 

Our proposed solution aims to support heterogeneous model architectures and to empower vendors to retain ownership of their model architectures, while enabling the composition and interoperability of modular blocks from different vendors.
\section{Problem Formulation and Proposed Scheme} 
\label{prob_formulation}
Building on the proposed architecture introduced in Section \ref{sys_model}, we now formalize the challenges in existing distributed learning paradigms and introduce our proposed solution. 
\subsection{Problem Formulation}
Formally, all clients collaboratively optimize a global model $\boldsymbol{\theta}$ by minimizing the weighted sum of their local empirical risks:
\begin{align} 
	\min_{\boldsymbol{\theta} \in \boldsymbol{\Theta}}\,\,  F({\boldsymbol{\theta}}) =  \displaystyle\sum_{k=1}^{N}\frac{d_k}{d} F_k(\boldsymbol{\theta}),
\end{align} 
where $d = \sum_{k=1}^N d_k$ is the total number of samples across all clients and $d_k$ is the number of samples at client $k$. The local objective of client $k$ is given by 
\begin{align}
F_k(\boldsymbol{\theta})=\mathbb{E}_{(x_k,y_k) \sim \mathcal{D}_k}[\ell(f(x_k;\boldsymbol{\theta}),y_k)] ,
\end{align}
where $(x_k,y_k)$ denotes the input-label pairs drawn from its dataset $\mathcal{D}_k$, $\ell$ is the loss function, and $\boldsymbol{\Theta}$ is the feasible set of model parameters. 

During the communication round $t$, the server sends the global model $\boldsymbol{\theta}^t$ to each client $k$, which then performs $\tau$ local Stochastic Gradient Descent (SGD) steps as follows \cite{ruder2017overviewgradientdescentoptimization}:
\begin{align}
	\boldsymbol{\theta}_k^{t,s} =  \boldsymbol{\theta}_k^{t,s-1} - \eta \nabla F_k(\boldsymbol{\theta}_k^{t,s-1}), \quad \forall s \in \{1,2, \cdots, \tau\},
\end{align} 
with $\boldsymbol{\theta}_k^{t,0} = \boldsymbol{\theta}^{t}$. Then, the clients send their local models to the server for aggregation:
\begin{align} 
	\boldsymbol{\theta}^{t+1} = \displaystyle\sum_{k=1}^N \frac{d_k}{d}\boldsymbol{\theta}_k^{t, \tau}.
\end{align} 
However, transmitting the full model parameters $\boldsymbol{\theta}_k^{t, \tau}$ to the server incurs high communication overhead, in addition to composing a global model $\boldsymbol{\theta}^{t+1}$ that must be shared among clients, undermining customization and ownership.

To reduce per‐round upload sizes, FSL enables partitioning the model at the cut-layer into a client‐side model $f_c(x_k;\boldsymbol{\theta}_{c,k})$, with client-specific parameters $\boldsymbol{\theta}_{c,k}$, and a server-side model $f_s(h_k;\boldsymbol{\theta}_s)$, with shared server parameters $\boldsymbol{\theta}_s$, where
\begin{align}
	 h_k = f_c(x_k; \boldsymbol{\theta}_{c,k})
\end{align} 
denotes the output of the cut-layer that is sent from client $k$ to the server. The server then continues the forward propagation to generate the predicted labels as follows:
\begin{align} 
	 \tilde{y}_k = f_s(h_k; \boldsymbol{\theta}_s).
\end{align} 
Transmitting the output of the cut-layer yields a smaller communication cost per round. Nonetheless, FSL allows the server to retain part of the model, with access to both the architecture and the parameters. Moreover, local end-to-end inference is not supported, as the server-side model is required for all clients, which hinders interoperability between clients. 

\subsection{Proposed Scheme}

We introduce a novel interoperable Federated Learning (IFL) approach that enables clients to retain their full model architecture and parameters locally while collaboratively training across heterogeneous models.
Each client $k$ decomposes its model into a base block $f_{b,k}(\cdot;\boldsymbol{\theta}_{b,k})$ and a modular block $f_{m,k}(\cdot;\boldsymbol{\theta}_{m,k})$, where $\boldsymbol{\theta}_{b,k}$ and $\boldsymbol{\theta}_{m,k}$ denote the parameters of the respective blocks. These blocks are connected by a fusion-layer whose output is denoted as $z_k$. Crucially, only the fusion-layer output $z_k$ (intermediate representation) and the corresponding labels $y_k$ are shared with the server.

Our training process alternates between updating the base blocks using local data samples and updating modular blocks using concatenated fusion‐layer outputs from all clients. This approach decouples personalization (base block) from generalization (modular block), while preserving model ownership and reducing communication costs. In what follows, we describe the update details of both blocks during communication round $t \in \{1,2,\ldots,T\}$. The detailed pseudocode can be found in Algorithm \ref{pseudocode}. 
\subsubsection{Base Block Update} 

Each client $k$ performs forward and backward propagation over a sequence of $\tau$ mini-batches drawn from its local dataset $\mathcal{D}_k$. During this local update phase, only the parameters of the base block are optimized via SGD, while the modular block remains fixed. The base block parameters $\boldsymbol{\theta}_{b,k}$ are updated according to
\begin{align} 
	 \boldsymbol{\theta}_{\text{b},k}^{t+1} = \boldsymbol{\theta}_{\text{b},k}^{t} - \eta_\text{b} \displaystyle\sum_{s=1}^{\tau} \Delta_{\boldsymbol{\theta}_{\text{b},k}}\ell(f_{\text{m},k}(f_{\text{b},k}(x_k^s;\boldsymbol{\theta}_{\text{b},k}^{t});\boldsymbol{\theta}_{\text{m},k}^{t}),y_k^s), 
\end{align}
where $\eta_\text{b}$ is the learning rate,  $\Delta_{\boldsymbol{\theta}_{\text{b},k}}\ell(\cdot)$ denotes the gradient of the loss with respect to the base block parameters, and $(x_k^s, y_k^s)$ represents the $s^{th}$ mini-batch sampled from $\mathcal{D}_k$.
\begin{algorithm}[H]
\caption{Interoperable Federated Learning (IFL)}
\label{pseudocode}
\begin{algorithmic}[1]
\STATE \textbf{Initialize:} $\boldsymbol{\theta}_{\text{b},k}^{0}$, $\boldsymbol{\theta}_{\text{m},k}^{0}$ for each client $k = 1$ to $N$
\FOR{each communication round $t = 0$ to $T-1$}
    \STATE \textbf{Base Block Update}
    \FOR{each client $k = 1$ to $N$ \textbf{in parallel}}
        \FOR{each local step $s = 1$ to $\tau$}
            \STATE Sample mini-batch $(x_k, y_k)$ from local dataset $\mathcal{D}_k$

            \STATE $\hat{y}_k \leftarrow f_{\text{m},k}(f_{\text{b},k}(x_k; \boldsymbol{\theta}_{\text{b},k}^{t});\boldsymbol{\theta}_{\text{m},k}^{t})$
            \STATE $\ell_k \leftarrow \text{Loss}(\hat{y}_k, y_k)$
            \STATE $\boldsymbol{\theta}_{\text{b},k}^{t} \leftarrow \boldsymbol{\theta}_{\text{b},k}^{t} - \eta_b \nabla_{\boldsymbol{\theta}_{\text{b},k}} \ell_k$
        \ENDFOR
    \STATE $\boldsymbol{\theta}_{\text{b},k}^{t+1} \leftarrow \boldsymbol{\theta}_{\text{b},k}^{t}$

    \ENDFOR

    \STATE \textbf{Fusion-Layer Output Transmission}
    \FOR{each client $k = 1$ to $N$ \textbf{in parallel}}
        \STATE Sample fresh mini-batch $(x_k, y_k)$
        \STATE $z_k \leftarrow f_{\text{b},k}(x_k; \boldsymbol{\theta}_{\text{b},k}^{t+1})$
        \STATE Send $(z_k, y_k)$ to server
    \ENDFOR
    \STATE \textbf{Server Concatenation and Broadcast}
    \STATE Server collects $Z = \{z_1, \dots, z_N\}$ and $Y = \{y_1, \dots, y_N\}$
    \STATE Server broadcasts $(Z, Y)$ to all clients
    \STATE \textbf{Modular Block Update}
    \FOR{each client $k = 1$ to $N$ \textbf{in parallel}}
        \FOR{each $i = 1$ to $N$}
            \STATE $\hat{y}_i \leftarrow f_{\text{m},k}(z_i; \boldsymbol{\theta}_{\text{m},k}^{t})$
            \STATE $\ell_i \leftarrow \text{Loss}(\hat{y}_i, y_i)$
            \STATE $\boldsymbol{\theta}_{\text{m},k}^{t} \leftarrow \boldsymbol{\theta}_{\text{m},k}^{t} - \eta_m \nabla_{\boldsymbol{\theta}_{\text{m},k}} \ell_i$
        \ENDFOR
        \STATE  $\boldsymbol{\theta}_{\text{m},k}^{t+1} \leftarrow \boldsymbol{\theta}_{\text{m},k}^{t}$
    \ENDFOR
\ENDFOR
\end{algorithmic}
\end{algorithm}
\begin{table}[!ht]
\caption{Comparison of key features across FL, FSL, and the proposed IFL framework}
\label{tab:comparison}
\centering
\begin{tabular}{|l|c|c|c|}
\hline
\textbf{Feature} & \textbf{FL} & \textbf{FSL} & \textbf{IFL} \\
\hline
Client parameters remain private   & \xmark & \cmark & \cmark \\
\hline
Local end-to-end inference         & \cmark & \xmark & \cmark \\
\hline
Lightweight uplink update          & \xmark & \cmark & \cmark \\
\hline
Multiple updates per round         & \cmark & \xmark & \cmark \\
\hline
Full model architecture privacy    & \xmark & \xmark & \cmark \\
\hline
Heterogeneous model support        & \xmark & \xmark & \cmark \\
\hline
Cross-client model composition      & \xmark & \xmark & \cmark \\
\hline
\end{tabular}
\end{table}
After the update of the base block parameters, each client $k$ samples a fresh mini‐batch $(x_k,y_k)$ of size $B$ and performs forward propagation on the base block as follows
\begin{align} 
	z_k = f_{\text{b},k}(x_k; \boldsymbol{\theta}_{\text{b},k}^{t+1}).
\end{align}

The client then sends the pair $(z_k, y_k)$ to the server, which concatenates all received data from the clients into $Z = \{z_1,z_2, \ldots, z_N\}$ and $Y = \{y_1, y_2, \ldots, y_N\}$, and broadcasts the pair $(Z,Y)$ to all the clients.
\subsubsection{Modular Block Update} 
Upon receiving the concatenated data from the server, each client $k$ performs a local update on its modular block parameters $\boldsymbol{\theta}_{\text{m},k}^t$, as follows
\begin{align} 
	 \boldsymbol{\theta}_{\text{m},k}^{t+1} = \boldsymbol{\theta}_{\text{m},k}^{t} - \eta_\text{m} \displaystyle\sum_{i=1}^N \Delta_{\boldsymbol{\theta}_{\text{m},k}}\ell(f_{\text{m},k}(z_i;\boldsymbol{\theta}_{\text{m},k}^{t}),y_i),
\end{align} 
where $\eta_m$ is the learning rate and $\Delta_{\boldsymbol{\theta}_{\text{m},k}}\ell(\cdot)$ is the gradient of the loss with respect to the modular block parameters.

During the inference phase, each client $k$  uses its local model to generate predictions $\hat{y}_k$ as follows
\begin{align} 
	\hat{y}_k = f_{\text{m},k}(f_{\text{b},k}(x_k;\boldsymbol{\theta}_{\text{b},k});\boldsymbol{\theta}_{\text{m},k}).
\end{align} 
Since modular blocks from any client can be composed with the base block of client $k$, interoperable inference can be performed as
\begin{align} 
	\hat{y}_{k,i} = f_{\text{m},i}(f_{\text{b},k}(x_k;\boldsymbol{\theta}_{\text{b},k});\boldsymbol{\theta}_{\text{m},i}), \quad \forall i=\{1,2,\ldots,N\},
\end{align} 
where $\hat{y}_{k,i}$ denotes the prediction for client $k$ using its own base block combined with the modular block of client $i$. Table~\ref{tab:comparison} summarizes the key differences between our proposed IFL architecture and existing schemes such as FL and FSL.
\vspace{-0.1cm}

\section{Numerical Results} \label{numerical}
\renewcommand{\arraystretch}{1.3}
\begin{table*}[ht]
\caption{Client-specific model architectures used in the IFL framework. The fusion-layer is highlighted in bold}
\label{tab:model_architectures}
\centering
\renewcommand{\arraystretch}{1.3}
\begin{tabular}{|c|l|l|}
\hline
\textbf{Client} & \textbf{Base Block Layers} & \textbf{Modular Block Layers} \\
\hline
1 & Conv(1,16) $\to$ Conv(16,32) $\to$ \textbf{Conv(32,48)} &
FC(432,256) $\to$ FC(256,128) $\to$ FC(128,64) $\to$ FC(64,10) \\
\hline
2 & Conv(1,16) $\to$ Conv(16,32) $\to$ \textbf{FC(1568,432)} &
FC(432,128) $\to$ FC(128,10) \\
\hline
3 & \textbf{FC(784,432)} &
FC(432,256) $\to$ FC(256,128) $\to$ FC(128,64) $\to$ FC(64,10) \\
\hline
4 & FC(784,1024) $\to$ FC(1024,512) $\to$ \textbf{FC(512,432)} &
FC(432,10) \\
\hline
\end{tabular}
\end{table*}

\subsection{Experimental Setup}
We consider a collaborative training setup with $N = 4$ clients and a central server. Each client performs $\tau = 10$ local iterations to update its base block before communicating with the server. The training is carried out for $T = 200$ communication rounds. We report results averaged over 10 Monte Carlo runs.
\subsubsection{Baselines}
Our proposed algorithm IFL is compared to the following baselines.
\begin{itemize}
    \item FL (Federated Learning): Clients train their local models on local data and transmit the updated models to the server, which aggregates them using FedAvg \cite{FL}, then sends the updated global model back to the clients.
    \item FSL (Federated Split Learning): Based on \cite{10118601}, clients share cut-layer outputs with the server. The server-side model updates are averaged, while client-side models remain personalized to local data.
 
\end{itemize}
\subsubsection{Dataset}
We consider an image classification task using Kuzushiji-MNIST dataset, where we have $50k$ training samples and $10k$ test samples \cite{japanese}. To measure the effect of data heterogeneity, we generate Non-Independent and Identically Distributed (Non-IID) data using a Dirichlet distribution with concentration parameter $\alpha$, which controls the per-class sample distribution among clients \cite{zipf_dist_paper}. We use the following configuration parameters: $\alpha$ = 0.5, $\eta_b = 0.01$, $\eta_m = 0.01$, $B= 32$, and $z_k \in \mathbb{R}^{B \times 432}, \quad \forall k \in \{1, 2, \ldots, N\}$. 
\subsubsection{Models}
To validate collaborative learning with heterogeneous models, each client is assigned a different architecture, reflecting vendor-specific implementations. Details of the models are summarized in Table \ref{tab:model_architectures}. Convolutional layers (Conv) are followed by pooling layers, while fully connected (FC) layers are followed by ReLU activations, except for the output layer. Importantly, the fusion-layer type may differ across clients, e.g., client 1 may use Conv-based fusion-layer, while others use FC-based layers. This highlights the importance of a standardized fusion-layer output dimension, without requiring identical fusion-layer types.
\begin{figure}[h]
	\centering
	\includegraphics[width = 0.48\textwidth]{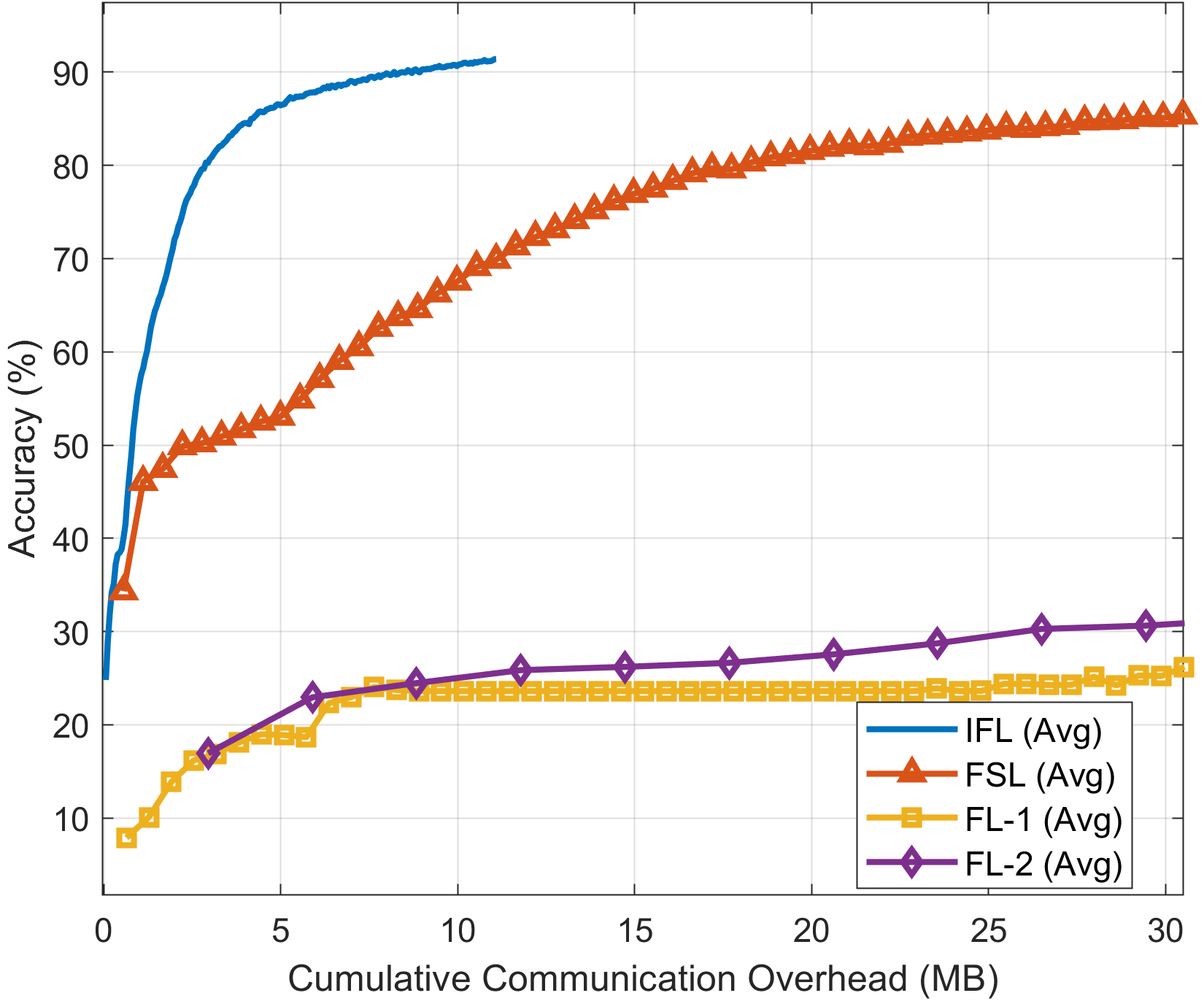}
	\caption{Test accuracy averaged over all clients versus cumulative communication overhead (MB) for IFL (proposed), FSL, and FL variants.} 
	\label{Fig:comm_eff} 
\end{figure}
\begin{figure}[h]
	\centering
	\includegraphics[width = 0.5\textwidth]{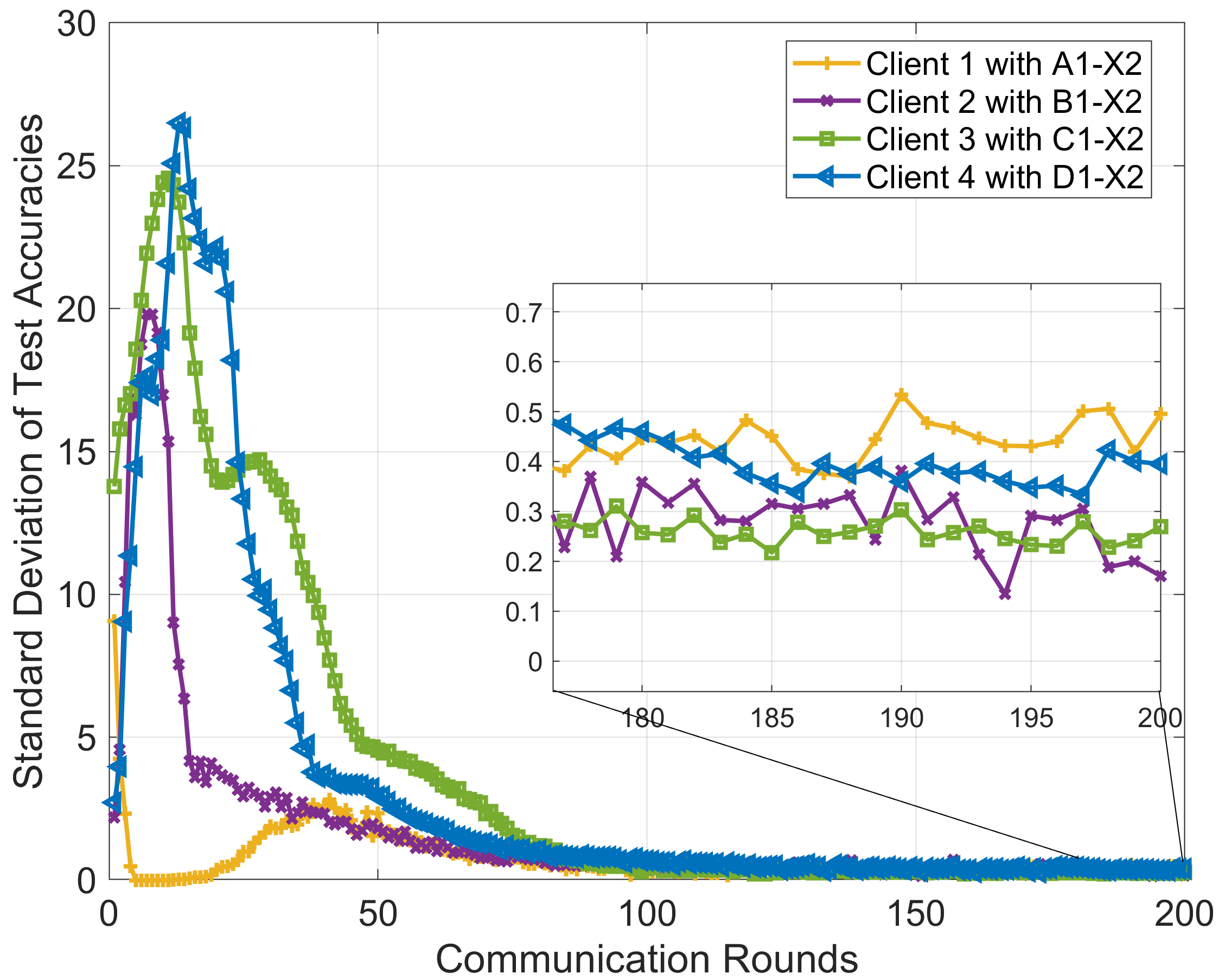}
	\caption{Standard deviation of test accuracy for each client’s base block combined with modular blocks from all clients over communication rounds.} 
	\label{Fig:std} 
\end{figure}
\begin{figure}[h]
	\centering
	\includegraphics[width = 0.48\textwidth]{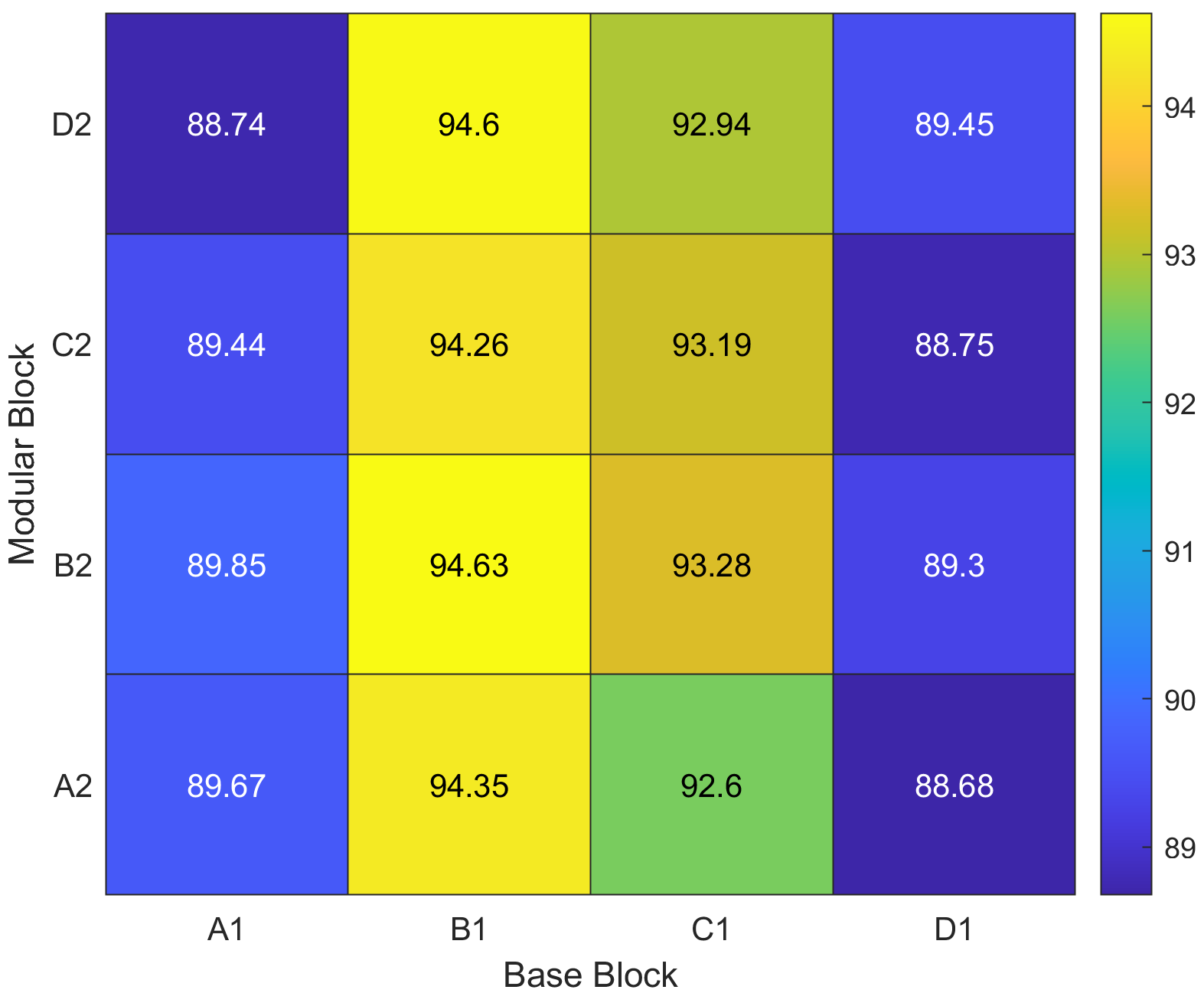}
	\caption{Test accuracy matrix for all base blocks and modular blocks combinations across clients.} 
	\label{Fig:heatmap} 
\end{figure}
\subsection{Results and Discussion}
\subsubsection{Communication Efficiency}
Fig. \ref{Fig:comm_eff} compares test accuracy versus cumulative communication overhead (in MB) incurred during training for IFL, FSL, and two FL variants. FL-1 deploys the smallest model (from client 1) across all clients, whereas FL-2 adopts the more accurate but larger model (from client 2). IFL achieves 90\% test accuracy with only 8.5~MB of uplink data, while FSL reaches just 64\% at the same communication cost. This demonstrates the advantage of IFL’s multiple local updates in reducing transmission frequency. Both FL variants incur high overhead due to transmitting full models in every communication round.
\subsubsection{Training with Heterogeneous Models}
Fig.~\ref{Fig:std} presents the standard deviation (SD) of the test accuracy for each client’s base block when paired with the modular blocks of all other clients over communication rounds. For example, the label $A1$–$X2$ denotes that client 1's base block ($A1$) is composed with all modular blocks from all the clients. At the end of the training, all SD values fell below 0.6, indicating consistent performance across heterogeneous combinations. This robustness was due to the fact that all modular blocks were trained on the same data sent by the server, which facilitated convergence despite architectural differences. These results confirmed IFL's effectiveness in supporting cross-vendor collaborative learning.
\subsubsection{Cross-Vendor Interoperable Inference}
We evaluate inference interoperability by constructing an accuracy matrix, where each entry shows the test accuracy from combining a client’s base block with another client’s modular block. Fig.~\ref{Fig:heatmap} shows that cross-client combinations achieve performance that is comparable to, and in some cases even exceed, that of local compositions (for instance, $A1$-$B2$ outperforms $A1$-$A2$, and $C1$-$B2$ surpasses $C1$-$C2$). This demonstrates the generalization and composability of the proposed framework.\vspace{-0.4cm}

\section{Conclusion} \label{conclusion}
We presented Interoperable Federated Learning (IFL), a communication-efficient framework for collaborative learning across clients with heterogeneous model architectures. By standardizing the fusion-layer output and decoupling model components, IFL enables cross-vendor training and modular inference without exposing model parameters or architectures. Experimental results demonstrate that IFL maintains robust performance across diverse model architectures and enables the seamless interchange of modular blocks across clients, highlighting its potential for scalable, interoperable deployment in multi-vendor AI systems.
\bibliographystyle{IEEEtran}
\bibliography{references}
\end{document}